\documentclass[letterpaper, 10 pt, conference]{ieeeconf}
\usepackage{times}
\usepackage{graphicx}
\usepackage{amsmath,amssymb,amsopn,amstext,amsfonts}
\usepackage{cancel}
\usepackage[space]{cite}
\usepackage{pdfsync}
\usepackage{balance}
\usepackage{color}
\usepackage{mathtools}
\usepackage{algpseudocode}
\usepackage{algorithm} 
\usepackage{bm}

\usepackage{diagbox}
\usepackage{float}
\usepackage{epstopdf}
\usepackage{pifont}
\usepackage{fixltx2e}
\usepackage{amsmath}
\usepackage{multirow}
\usepackage{url}
\usepackage{verbatim}
\makeatletter
\let\NAT@parse\undefined
\makeatother
\usepackage[linkcolor=black,citecolor=black,urlcolor=black,colorlinks=TRUE]{hyperref}
\usepackage{booktabs}
\usepackage{graphicx}
\usepackage{subcaption}
\usepackage{stfloats}
\newcommand{\tabincell}[2]{\begin{tabular}{@{}#1@{}}#2\end{tabular}}  

\graphicspath{{./}}
\DeclareGraphicsExtensions{.png,.jpg,.eps,.pdf}
\IEEEoverridecommandlockouts
	\title{\LARGE \bf Visibility-aware Trajectory Optimization 
		
		with Application to Aerial Tracking}
	\author{Qianhao Wang, Yuman Gao, Jialin Ji, Chao Xu, and Fei Gao
		\thanks{
			All authors are with the State Key Laboratory of Industrial Control Technology, Zhejiang University, Hangzhou 310027, China, and also with the Huzhou Institute of Zhejiang University, HuZhou 313000, China. E-mails:\tt\small \{qhwangaa, ymgao, jlji, cxu, fgaoaa\}@zju.edu.cn}
	}

\begin{document}
	
	\maketitle
	\thispagestyle{empty}
	\pagestyle{empty}
	\begin{abstract}
	The visibility of targets determines performance and even success rate of various applications, such as active slam, exploration, and target tracking.
	Therefore, it is crucial to take the visibility of targets into explicit account in trajectory planning. 
	In this paper, we propose a general metric for target visibility, considering observation distance and angle as well as occlusion effect. 
	We formulate this metric into a differentiable visibility cost function, with which spatial trajectory and yaw can be jointly optimized.
	Furthermore, this visibility-aware trajectory optimization handles dynamic feasibility of position and yaw simultaneously. 
	To validate that our method is practical and generic, we integrate it into a customized quadrotor tracking system. 
	The experimental results show that our visibility-aware planner performs more robustly and observes targets better. 
	In order to benefit related researches, we release our code to the public.

	\end{abstract}
	
	\IEEEpeerreviewmaketitle
	\section{Introduction}
	\label{sec:introduction}
	
	In recent years, the progress on various aspects of autonomous robots makes it possible to accomplish complex, systematic tasks. When executing tasks, robots usually equip sensors to obtain external environment information. However, the sensors usually have a limited field of view (FOV), especially for quadrotors under the size, weight, and power (SWaP) constraints. Therefore, the robot's position and orientation should be adjusted to obtain better visibility to the target.

	Visibility is of vital importance in a variety of applications. We take three scenarios as instances: \textbf{1)} For visual-inertial state estimation, enhancing the visibility of co-visible features can increase the state estimation accuracy significantly\cite{murali2019perception, zhang2018perception}. \textbf{2)} For exploration tasks, the target's visibility determines the quality and confidence of obtained information \cite{ly2019autonomous, freitag2018interactive}. \textbf{3)} For tracking, keeping high visibility of target is the key to avoid target loss which is the main reason for mission failure\cite{zhou2020raptor,bandyopadhyay2009motion}. In conclusion, different circumstances put forward the same requirement: visibility-aware trajectory planning.
		
	However, only a few works take visibility into account in trajectory planning. Most of them use probabilistic or deterministic searching methods with visibility cost to deal with the problem. Whereas these methods suffer from inherent inaccuracy caused by discretization. Besides, when the dimension of state increases, the computational complexity rises exponentially. These adverse properties lead to planning failures, especially with a limited time budget. Additionally, yaw planning approaches for visibility are usually absent. A group of works directly set facing-target yaw as expectation input in control level \cite{BJ2020ICRA, JC2016tracking}. However, in practice, the robot's yaw rate, named as yaw feasibility in this paper, can not be changed suddenly, same as the velocity and acceleration of position. (The feasibility mentioned in the following includes yaw feasibility by default) The above control-level strategy fails to reach the expected yaw sometimes due to physical limits. Some other works plan yaw after position, which omits the coupling relationship between them when considering visibility. To summarize, an efficient trajectory planning method that takes visibility into consideration comprehensively is rare.

	\begin{figure}[t]
		\vspace{0.2cm}
		\centering
		\includegraphics[width=1\linewidth]{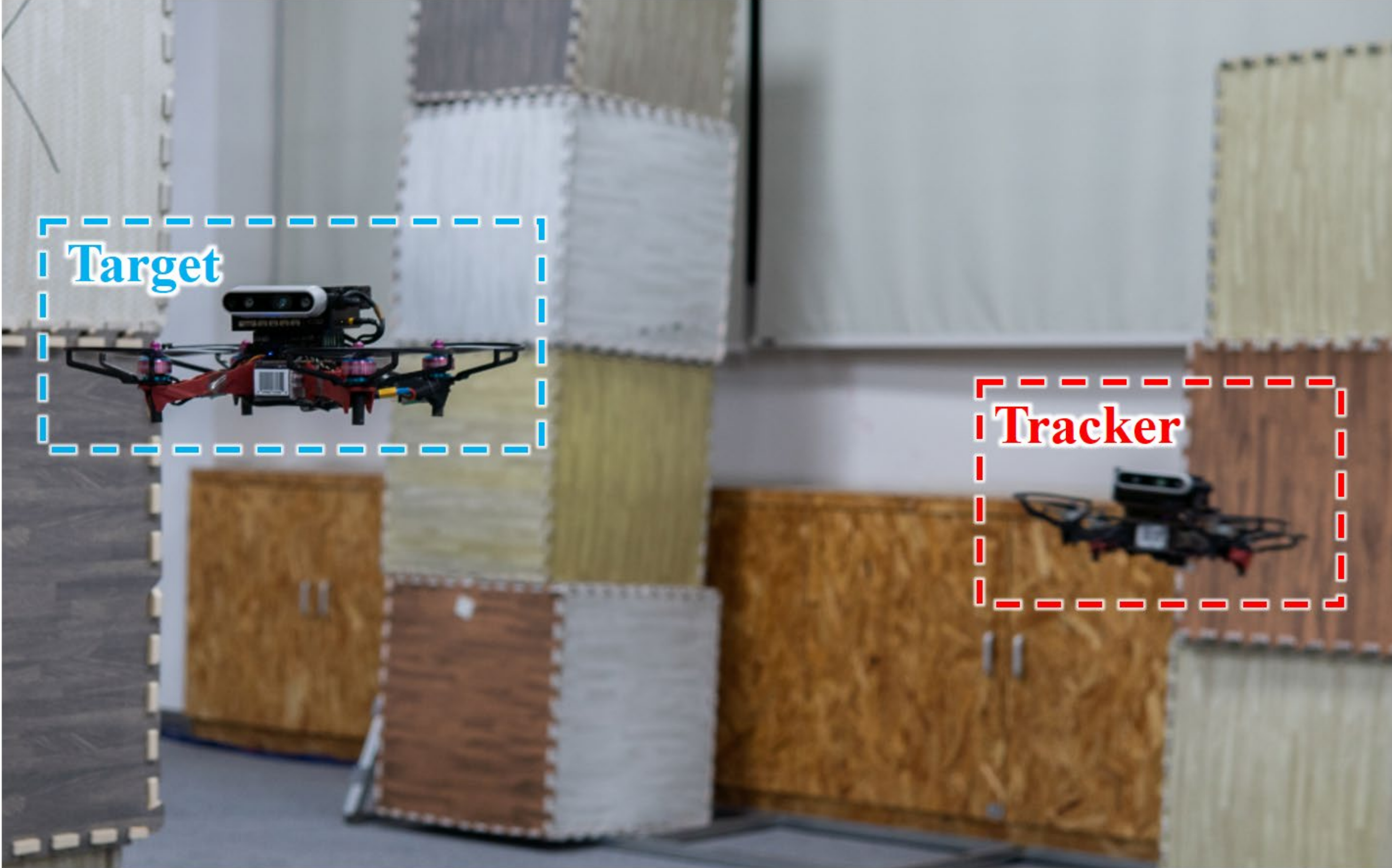}
		\captionsetup{font={small}}
		\caption{
			The real-world experiment of the aerial tracking application using our method.
		}
		\label{pic:tou2}
		\vspace{-0.6cm}
	\end{figure}
	
	\begin{figure}[t]
		\vspace{0.0cm}
		\centering
		\includegraphics[width=1\linewidth]{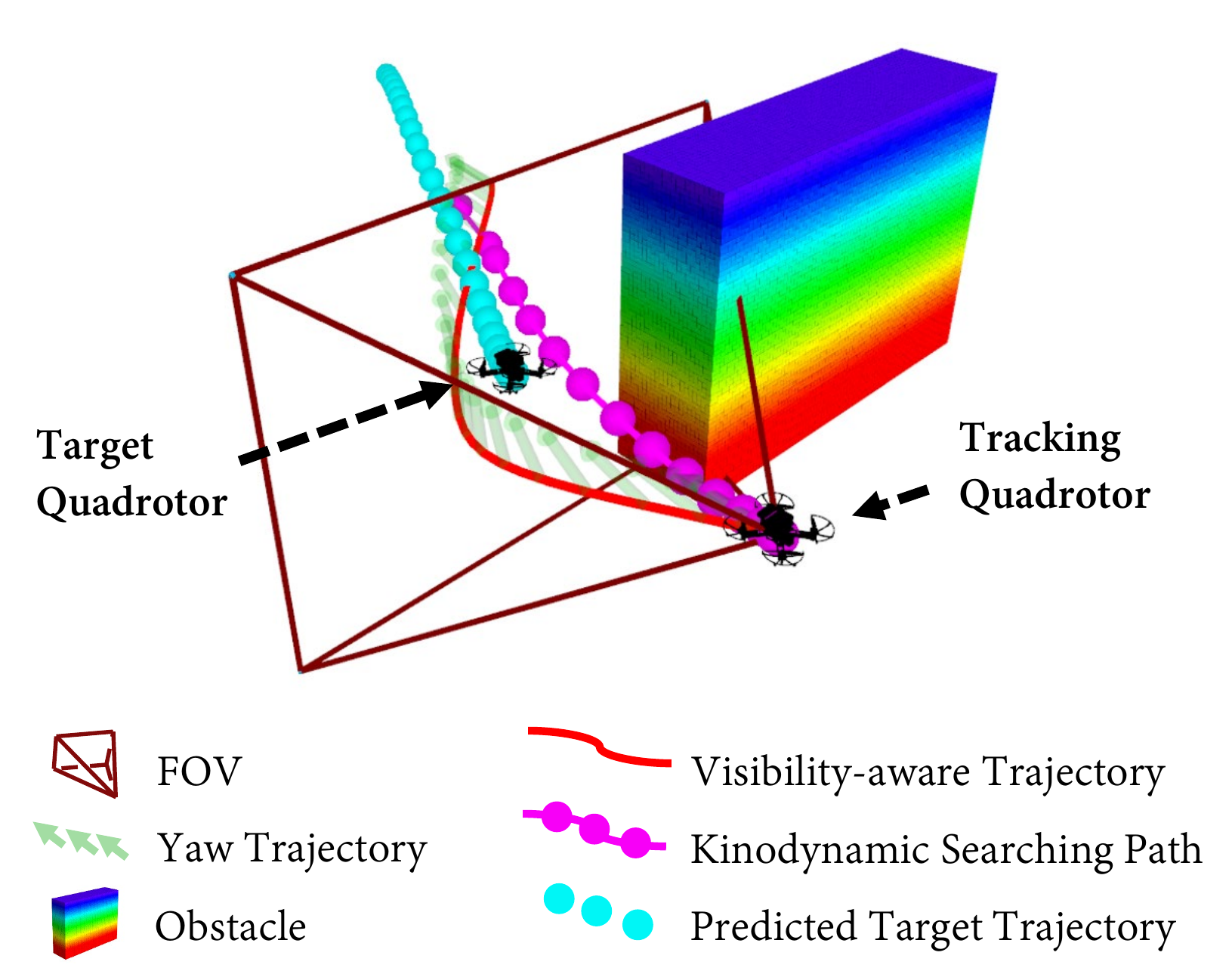}
		\captionsetup{font={small}}
		\caption{
			Illustration of the visibility-aware trajectory result in the application of aerial tracking.
		}
		\label{pic:tou}
		\vspace{-1.3cm}
	\end{figure}

\begin{figure}[t]
		\centering
		\includegraphics[width=1\linewidth]{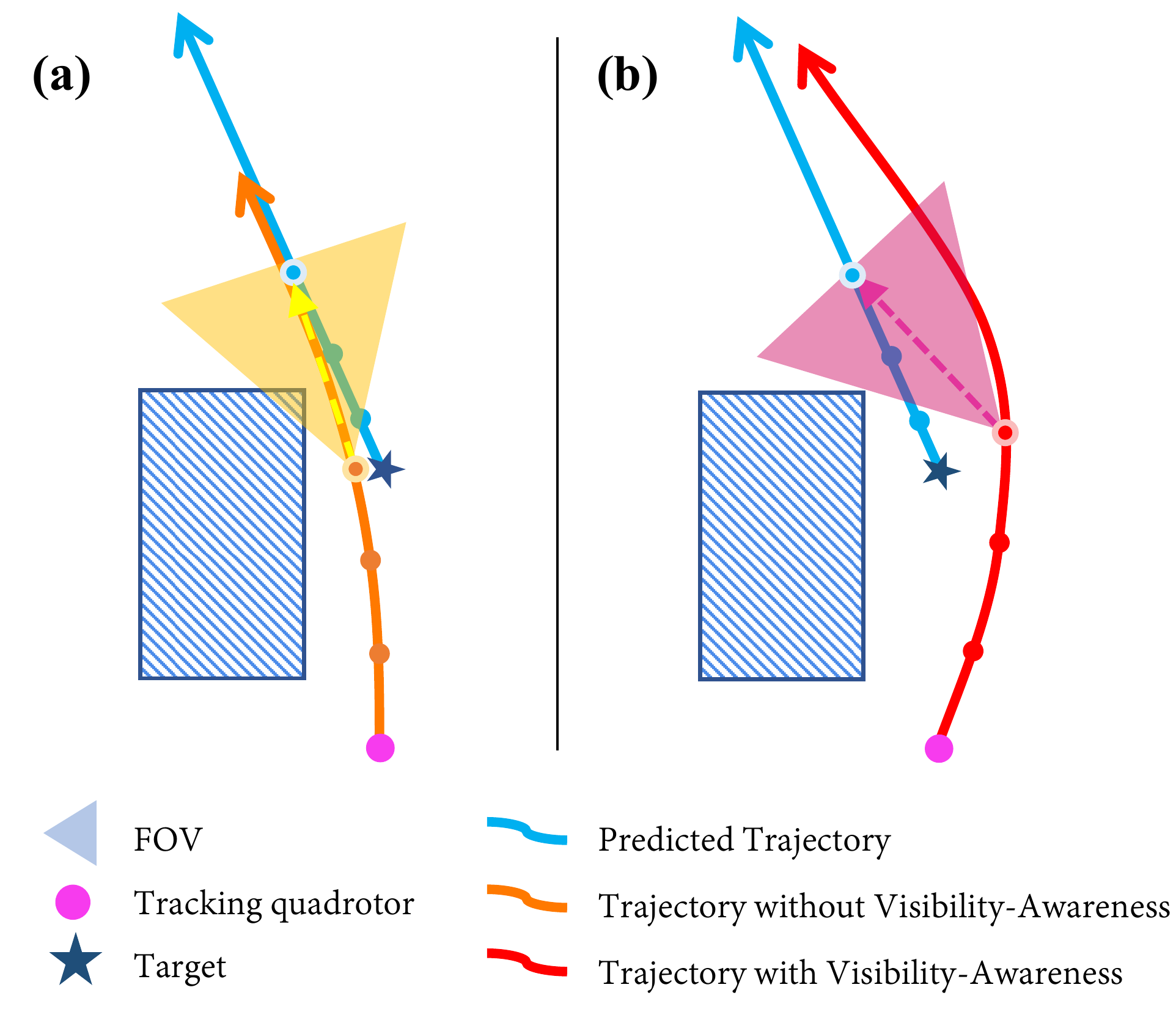}
		\captionsetup{font={small}}
		\caption{
			Comparison of trajectories with and without visibility-awareness. The balls on the trajectory represent the position in time. Without considering visibility, the orange trajectory is easy to lose target because of occlusion, especially when the target makes movements that are hard to predict. (a): the trajectory without visibility-awareness. (b): the trajectory with visibility-awareness.
		}
		\label{pic:visibility_com}
		\vspace{-0.6cm}
	\end{figure}

	To bridge this gap, we propose a general optimization-based visibility-aware trajectory planning method. This method uses a continuous and differentiable polynomial that prevents dimension explosion when the planning problem scales up. Furthermore, our method optimizes yaw jointly with position considering visibility, safety, and feasibility. The visibility is defined for a specific target or an area of interest. According to practical experience, we summarize three parts of the visibility criteria:
	
	\begin{itemize}
		\item [a)]
		\textit{Distance of observation (DO)}: A moderate distance from target is expected.
		\item [b)]
		\textit{Angle of observation (AO)}: The axis of the sensor's FOV is expected to be straight towards the target.
		\item [c)]
		\textit{Occlusion effect against obstacles (OE)}: The trajectory prevents the line of sight towards the target from occlusion. Without considering \textit{OE}, occlusion is prone to happen, as in Fig.\ref{pic:visibility_com}, for instance.
	\end{itemize}	

	To provide a gradient of visibility in trajectory optimization, we design a differentiable visibility cost considering the above parts. To guarantee the joint optimality of position and observation angle, we design a joint optimizer that simultaneously optimizes position and yaw using the visibility cost, safety, and feasibility cost. The planned motions provide the robot with better visibility to a target, as Fig.\ref{pic:tou} is shown.
	
	Finally, to prove that our method is practical and generic, we integrate the proposed method into a customized quadrotor tracking system, with a kinodynamic occlusion-avoid searching front-end to generate an initial path. We compare the tracking system using our method with the cutting-edge tracking works. The experimental results show that our visibility-aware planner performs more robustly in tracking.

	The contributions of this paper are:
	\begin{itemize}
		\item [1)] 
		A general differentiable visibility cost taking the aforementioned \textit{DO}, \textit{AO} and \textit{OE} into account.
		\item [2)]
		Considering visibility, safety and dynamic feasibility, a joint trajectory optimizer optimizes position and yaw of a robot simultaneously.
		\item [3)]
		Simulations and real-world tests in aerial tracking application validate that our method is practical and generic. Moreover, we make our method open source \footnote{https://github.com/ZJU-FAST-Lab/visPlanner}.
		
	\end{itemize}

	\section{Related Work}
	\label{sec:related_works}
	The related works of trajectory planning to maximize visibility can be divided into two groups: local control based methods and trajectory generation based methods. 
	\subsection{Local Control Based Methods}
	
	Several previous works formulate the trajectory planning and control integrally as a local control problem. Penin et al. \cite{BP2018elli} replan the trajectory online considering the \textit{OE} constraint by solving an optimization problem in the image space. Merging re-active control and planning, this method directly obtains the optimal control command. Lacking the environmental perception, this approach requires complete knowledge of the environment, thus not applicable in real applications. Andersen et al. \cite{andersen2017trajectory} consider \textit{OE} in trajectory planning to get more information for vehicle overtaking. They generate motions by maximizing the visibility ahead of obstacles with an MPC receding horizon planner. However, this task-specific method is hard to extend to other scenarios. Nägeli et al. \cite{TN2017elli} also take \textit{OE} and collision simultaneously into account in an MPC planner. A modular cost function based on the re-projection error of targets is designed to account for visibility. While this method fails to maintain its safety against obstacles with arbitrary shape since it assumes that all the obstacles are shaped by ellipsoid.
	
	\subsection{Trajectory Generation Based Methods}
	Trajectory generation based methods take the visibility into account and guarantee the smoothness and feasibility simultaneously. Bonatti et al.\cite{Bonatti2018AutonomousDC} present a real-time covariant gradient descent method to trade-off smoothness, obstacle avoidance, and \textit{OE} in tracking trajectory planning. Their \textit{OE} cost requires the connection line between the target and the robot to be free of obstacles within a certain range. However, this \textit{OE} cost does not take the conical FOV shape of most sensors into account. Bandyopadhyay et al.\cite{bandyopadhyay2009motion} define the shortest distance that the target escapes from the chasing robot's visible region as a risk function for \textit{OE}. Nevertheless, the precise geometry of the robot's visible region is required. In contrast to\cite{bandyopadhyay2009motion}, several researches\cite{BJ2019IROS, BJ2020ICRA} design the minimum value of the Euclidean Signed Distance Field (ESDF) on the line between the target and the robot as visibility metric for \textit{OE}. However, because this metric is not differentiable, it is not capable of being used to optimize visibility directly. To deal with visibility, even if the metric is differentiable, Jeon et al. \cite{BJ2020ICRA} propose a bi-level tracking planner. They search a graph incorporating safety and \textit{OE} metric to obtain a chasing corridor used for trajectory generation. However, building and traversing a graph is time-consuming. Typically, even with a low speed (1.2m/s) and coarse grid map (0.5 m) resolution, this graph-search based method takes over 200 ms\cite{han2020fast}. With similar \textit{OE} metric, Zhou et al. \cite{zhou2020raptor} propose a risk-aware trajectory refining strategy. Instead of explicitly optimizing \textit{OE}, to achieve risk-awareness, they use a task-specific method that guarantees sufficient distance to stop if the trajectory passes over a previously unknown obstacle. However, we formulate our general visibility metric into a differentiable cost, enabling us to optimize visibility explicitly.
	
	Most works of trajectory optimization considering visibility only take \textit{OE} into account, while as mentioned in Sec.\ref{sec:introduction}, rare works plan a trajectory for \textit{AO} and \textit{DO}. Some works  \cite{BJ2020ICRA, JC2016tracking} greedily choose the yaw angle facing the target all the time, omitting the yaw feasibility. Other methods such as Zhou et al. \cite{zhou2020raptor} optimize the yaw trajectory that passes through a yaw sequence searched by a graph while never joints with the position. There is a strong mutual influence between position and yaw when considering visibility. Our method considers \textit{AO}, with which we are capable of optimizing yaw with position jointly.

	\section{Preliminaries}
	\label{sec:Preliminaries}
	\subsection{Problem Statement}
	\label{sec:bezier_prediction}
	Given the sequence of a chosen target position in time, the joint optimizer's goal is to generate a trajectory that maximizes the target's visibility, guaranteeing safety and dynamic feasibility. Here the visibility is defined as the three parts above: \textit{DO}, \textit{AO} and \textit{OE}.
	
	We consider the problem based on the following assumptions:
	
	\begin{itemize}
	\item [1)]
	Only one target can be chosen at the same time.
	\item [2)] 
	The target is static or has smooth and bounded velocity and acceleration.
	\item [3)]
	The sensor equipped directly in front of robot has conical limited field of view.
	\end{itemize}	

	\subsection{Analytical Visibility Metric}
	Notating the position of the robot $\mathbf{p}=[x_p,y_p,z_p] \in \mathbb{R}^{3}$, Euler-yaw angle of the robot $\psi_p \in SO(2)$ and the position of the target $\mathbf{c}=[x_c,y_c,z_c] \in \mathbb{R}^{3}$, we formulate the details of the three parts of visiblity metric as follows.

	\subsubsection{DO}
	The target is expected to be observed in a proper distance range from the robot denoted by
	\begin{equation}\label{equ:do}
		od_{min} < ||{\mathbf{p}-\mathbf{c}}|| < od_{max},
	\end{equation}
	where the $od_{min}$ and $od_{max}$ are the lower and upper bounds of the optimal distance of observation. 

	\subsubsection{AO}
	In order to keep the axis of the sensor's FOV straight towards the target, the yaw angle is expected to be equal to $\psi_{best}$ defined as Eq.\ref{equ:ao}.
	\begin{equation}\label{equ:ao}
		\psi_{best} = {\rm atan2} \left(\boldsymbol e_y^T (\mathbf{p}-\mathbf{c}), \boldsymbol e_x^T (\mathbf{p}-\mathbf{c}) \right), 
	\end{equation}
	where $\boldsymbol e_x = [1, 0, 0]^T$ and $\boldsymbol e_y = [0,1,0]^T$.

	\subsubsection{OE}	
	\label{sec:OE}
\begin{figure}[t]
		\centering
		\includegraphics[width=1\linewidth]{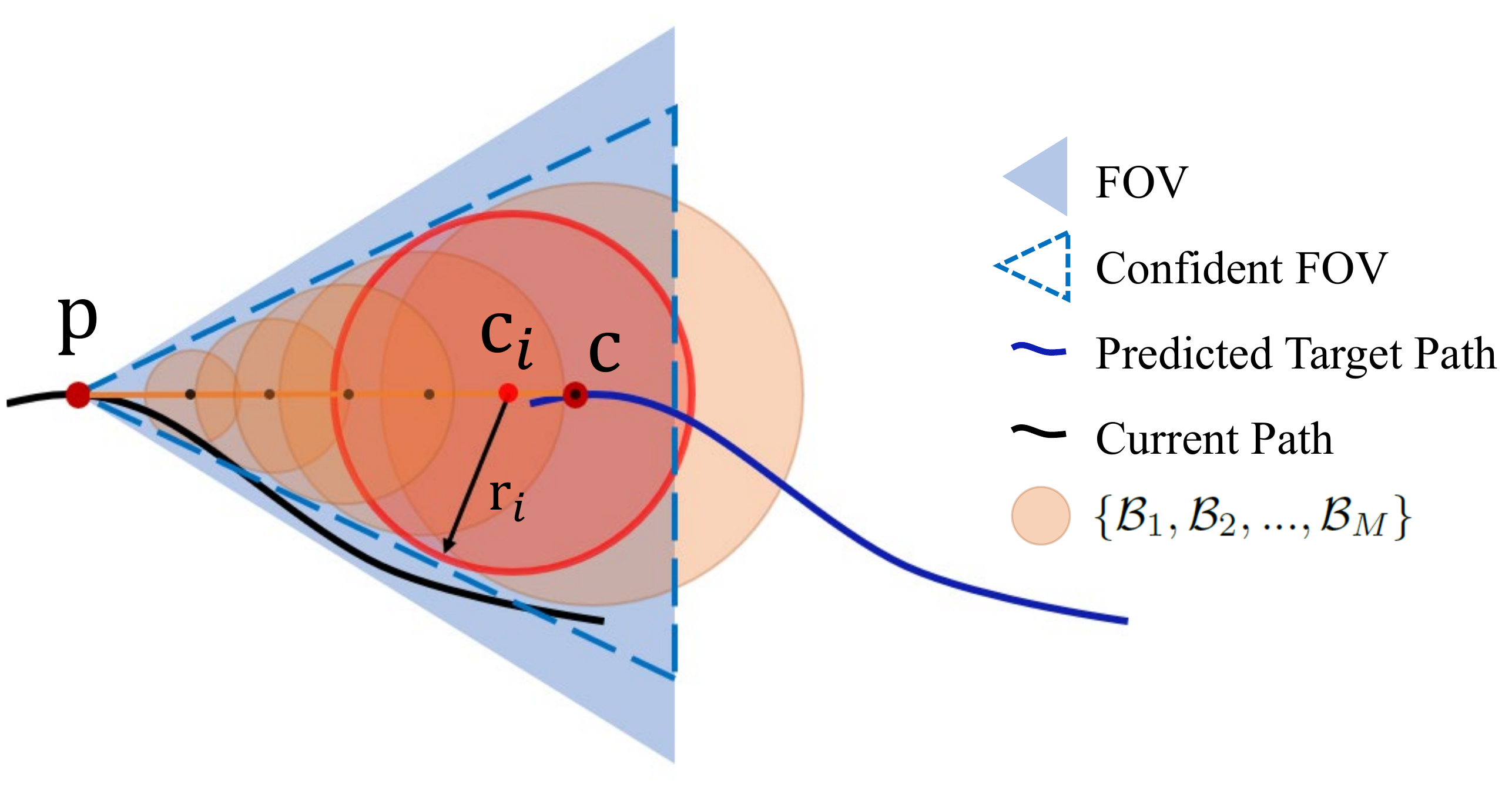}
		\captionsetup{font={small}}
		\caption{
			Illustration of the definition of \textit{OE} metric. A robot at position $\mathbf{p}$ observes the target $\mathbf{c}$. A sequence of ball-shaped areas in orange are used to approximate the confident FOV expected to be obstacle-free. 
		}
		\label{pic:visi_metric}
		\vspace{-0.5cm}
	\end{figure}

	As the two-dimension profile illustrates in Fig.\ref{pic:visi_metric}, the blue range represents the FOV of the robot. The blue dashed enveloped area represents the \textbf{confident FOV} which we expect the target within and no obstacles included. However, this requirement cannot be represented analytically. Therefore, we approximate the confident FOV with a sequence of ball-shaped areas $\{ \mathcal{B}_1,\mathcal{B}_2,...,\mathcal{B}_M \}$. A ball $\mathcal{B}_i$ is shown as the red circle with its center $\mathbf c_i$ and radius $r_i$ calculated by
	\begin{align}\label{eq:pi}
		\mathbf c_i &= \mathbf{p} + \lambda_i(\mathbf{c}-\mathbf{p}), \\
		r_i &= \rho \cdot \lambda_i \cdot ||{\mathbf{p}-\mathbf{c}}||, \label{eq:r_i}
	\end{align}
	where $\lambda_i = i/M \in [0, 1]$, and $\rho$ is a constant determined by the size of confident FOV.
	Then we can guarantee OE analytically for each ball by
	\begin{equation}
		\label{eq:oe}
		r_i < \Xi(\mathbf c_i), 
	\end{equation}
	where $\Xi(\mathbf c_i): \mathbb R^3 \rightarrow \mathbb{R}$ is the distance to the closest obstacle.
	
	\section{Visibility-aware Trajectory Optimization}

	\subsection{Trajectory Representation} 
	\label{sec:Trajectory Representation}
	We require a joint trajectory optimizer, which can generate a dynamic-feasible and collision-free trajectory considering smoothness and visibility simultaneously. 

	The trajectory is parameterized by a four-dimensional unclamped uniform B-spline curve, which is a piecewise polynomial uniquely determined by its degree \emph{p}$_b$, a knot span $\Delta t$, and \emph{N}$_c$ control points $\{\mathbf Q_i, \Phi_i\} \in \mathbb{R}^3 \times SO(2)$. 	
	We follow the work of [15] to plan the control points of positions $\mathbf Q_i$ and yaw angles $\Phi_i$ in a reduced space of the four selected differentially flat outputs $\{x,y,z,\psi\}$.

	\subsection{Objective Functions}
	In practice, we choose $p_b = 3$, then the total duration of the trajectory is $(N_c-3)\cdot \Delta t$. Since some of the objective functions are represented by the expressions of each waypoint $\{\mathbf p_{k}, \psi_{p,k}\}$ and distance $d_k$ to the corresponding target $\mathbf c_k$, they should be calculated in advance:
	\begin{align}\label{eq:waypoint}
		\mathbf{p_{k}} &= \frac{1}{6} (\mathbf{Q}_{k} + 4\mathbf{Q}_{k+1} + \mathbf{Q}_{k+2}),
	\end{align}
	\begin{align}
		d_k &= ||\mathbf p_k-\mathbf c_k||, k \in \{1,2,...,N_c-2\}.
	\end{align}
	Besides, $\psi_{p,k}$ can be calculated by yaw control point $\Phi_k$ similarly as Eq.\ref{eq:waypoint}.

	Additionally, we use a $C^2$ penalty function $g(x)=\max\{0, x\}^3$ to convert all constraints to penalty terms, and the final unconstrained optimization problem is given by
	\begin{equation}
		\min _{\mathrm{Q, \psi}} J= [J_{DO}, J_{AO}, J_{OE}, J_f, J_{f_\Phi} J_s, J_{s_\Phi}, J_c] \cdot \boldsymbol{\omega},
	\end{equation}
	where $\boldsymbol{\omega}$ is a weighing vector for trading off each cost:

	\subsubsection{DO Cost}
	To achieve proper observation distance, each waypoint $\mathbf{p_{k}}$ is required to satisfy \textit{DO} constraint defined by Eq.\ref{equ:do}.
	The \textit{DO} cost is written by:
	\begin{equation}
		J_{DO}=\sum_{k=1}^{N_c-2} g\left(  od_{min}^2-d_k^2 \right) + g\left( d_k^2 - od_{max}^2 \right).
	\end{equation}
	
	\subsubsection{AO Cost}
	The yaw angle is expected to be close to the best yaw $\psi_{best}$ defined by Eq.\ref{equ:ao}.
	The \textit{AO} cost is written by:
	\begin{equation}
		J_{AO} = \sum_{k=1}^{N_c-2}(\psi_{p,k} - \psi_{best,k})^2.
	\end{equation}
	Different from other costs, this gradient affects both position  $\mathbf {p_k}$ and yaw $\psi_{p,k}$:
	\begin{equation}
		\frac{\partial{J_{AO}}}{\partial{\psi_{p,k}}} = 2(\psi_{p,k} - \psi_{best,k}),
	\end{equation}
	\begin{equation}
		\frac{\partial{J_{AO}}}{\partial{\mathbf{p_k}}} = \left[\frac{\partial{J_{AO}}}{\partial{x_{p,k}}},\frac{\partial{J_{AO}}}{\partial{y_{p,k}}},0  \right]^T,
	\end{equation}
	where
	\begin{equation}
		\frac{\partial{J_{AO}}}{\partial{x_{p,k}}} = \frac{2(\psi_{p,k} - \psi_{best,k})}{(\boldsymbol e_x^T\mathbf{L})^2 + (\boldsymbol e_y^T\mathbf{L})^2} \cdot \boldsymbol e_y^T\mathbf{L},
	\end{equation}
	$\partial{J_{AO}}/\partial{y_{p,k}}$ can be calculated similarly. With Eq.\ref{eq:waypoint}, we can propagate the gradient back to control points.
	
	\subsubsection{OE Cost}
	To satisfy the requirement defined by Eq.\ref{eq:oe}, the \textit{OE} cost is written by:
	\begin{equation}
		J_{OE} =  \sum_{k=1}^{N_c-2}\sum_{i=1}^{M}g\left(F_{OE}(\mathbf{p_k})\right),
	\end{equation}
	\begin{equation}
		F_{OE}(\mathbf{p_k}) =  (r^2_{k,i} - \Xi^2(\mathbf c_{k,i})),
	\end{equation}
	where $\Xi(\mathbf c_{k,i})$ is obtained from ESDF, and $M$ is the number of ball-shaped areas mentioned in Sec.\ref{sec:OE}.
	
	The gradient of $F_{OE}$ cost can be written as:
	\begin{equation}
		\begin{aligned}
			\frac{\partial{F_{OE}}}{\partial{\mathbf{p_k}}} = & 2 
			\left[\rho r_{k,i}  \lambda_i \cdot \frac{\partial{d_k}}{\partial{\mathbf{p_{k}}}} 
			-(1-\lambda_i)\Xi(\mathbf c_{k,i}) \frac{\partial{\Xi(\mathbf{c}_{k,i})}}{\partial{\mathbf{c}_{k,i}}}
			\right],
		\end{aligned}
	\end{equation}
	where $\partial\Xi(\mathbf c_{k,i})/\partial\mathbf{c}_{k,i}$ is easily obtained from ESDF. The term $\lambda_i$ acts as a lever for the gradient.
	Then we use Eq.\ref{eq:waypoint} to propagate the gradient back to control points.

	\subsubsection{Dynamic Feasibility Cost}
	The control points of velocity $\textbf V_i$, acceleration $\textbf A_i$ and jerk $\textbf J_i$ are calculated by
	\begin{align}
		\textbf{V}_i = \frac{\textbf{Q}_{i+1} - \textbf{Q}_i}{\Delta t},  
		\textbf{A}_i = \frac{\textbf{V}_{i+1} - \textbf{V}_i}{\Delta t},
		\textbf{J}_i = \frac{\textbf{A}_{i+1} - \textbf{A}_i}{\Delta t}.
	\end{align}
	The control points of yaw velocity $V_{\Phi,i}$, yaw acceleration $A_{\Phi,i}$ and yaw jerk $J_{\Phi,i}$ can also be calculated by $\Phi$ similarly.
	Benefiting from the convex-hull property, dynamic feasibility of the whole trajectory can be guaranteed sufficiently by
	\begin{align}
		J_f = \sum_{i=1}^{N_c-1} g(||\textbf{V}_{i}||^2-v_{m}^2) + \sum_{i=1}^{N_c-2} g(||\textbf{A}_{i}||^2-a_{m}^2), \label{cost function}
	\end{align}
	where $v_{m}, a_{m}$ are the limits on velocity and acceleration.
	Similarly, the yaw feasibility cost can be written as:
	\begin{align}
		J_{f_\Phi} = \sum_{i=1}^{N_c-1} g(||V_{\Phi,i}||^2-v_{\Phi,m}^2) + \sum_{i=1}^{N_c-2} g(||A_{\Phi,i}||^2-a_{\Phi,m}^2),
	\end{align}
	where $v_{\Phi,m}, a_{\Phi,m}$ are the limits on velocity and acceleration of yaw.
	
	\subsubsection{Smoothness Cost}
	Benefiting from the convex hull property, we minimize the control points of high order derivatives of the B-spline trajectory to guarantee the smoothness:
	\begin{equation}
		J_{s}=\sum_{i=1}^{N_{c}-3}\left\|\mathbf{J}_{i}\right\|_{2}^{2},
	\end{equation}
	Similarly, we define smoothness cost of yaw $J_{s_\Phi}$ using $J_{\Phi,i}$.

	\subsubsection{Collision Cost}
	To guarantee that the entire trajectory is collision-free, the collision penalty serves as the repulsive force that pushes the control points away from obstacles.
	\begin{equation}
		J_{c}=\sum_{i=1}^{N} g(d^2_{t h r} - \Xi^2(\mathbf{Q}_{i}))\left(\Xi\left(\mathbf{Q}_{i}\right)\right),
	\end{equation}
	where $d_{thr}$ is the safety threshold distance.

	\begin{figure*}[t]
		\centering
		\includegraphics[width=1\linewidth]{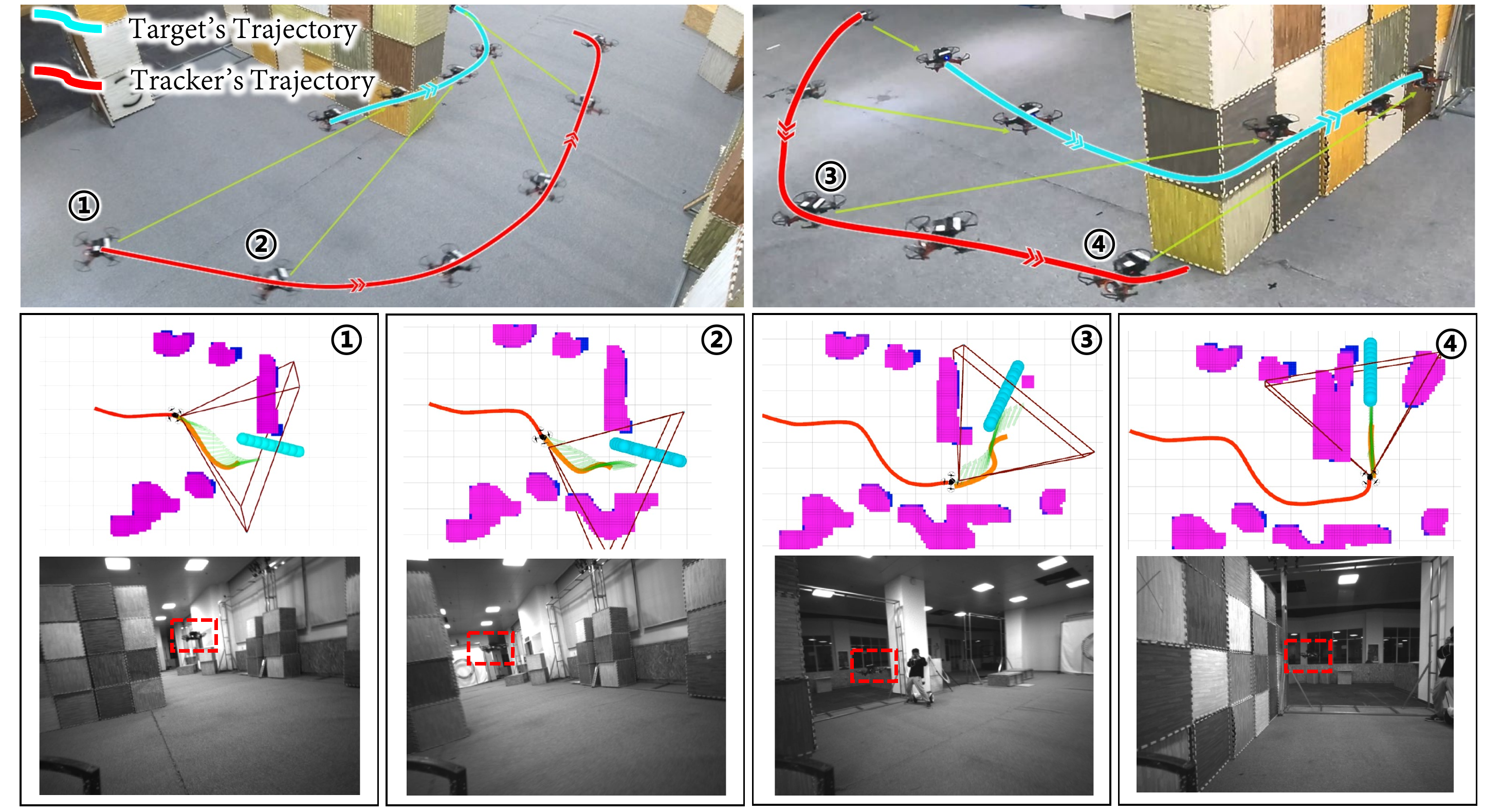}
		\captionsetup{font={small}}
		\caption{
			The real-world experiment. Four moments \ding{172}, \ding{173}, \ding{174}, \ding{175} are extracted from the entire experiment. \textbf{Top}: the snapshots of the tracking quadrotor and the target quadrotor. 
			\textbf{Middle}: the visualization of the algorithm and the experiment. For the tracking quadrotor, the red curve denotes the executed trajectory, the orange curve denotes the planned trajectory, and the green arrows represent the yaw trajectory. Blue balls depict the predicted trajectory of the target. The red pyramid stands for FOV. \textbf{Bottom}: the first-person perspective of the tracking quadrotor. The target is boxed out by the red dotted box.
		}
		\label{pic:snapshot11}
		\vspace{-0.3cm}
	\end{figure*}

	\section{Application on Quadrotor Tracking}
	\subsection{Motivation and Methodology}

	For tracking, keeping high visibility of target is critical to preventing target loss which is the main reason for failures. Although with a requirement of visibility in the existing tracking systems, they rarely consider the complete metrics we summarize through practical experiments. To validate our method, we integrate the proposed optimizer into our previous work\cite{han2020fast}, a customized quadrotor tracking system. Its back-end is replaced by the proposed method to generate trajectory considering visibility and yaw feasibility. The predicted target position is generated by the same method based on Bézier regression in our previous work.

	\begin{figure}[t]
		\vspace{-0.0cm}
		\centering
		\includegraphics[width=1\linewidth]{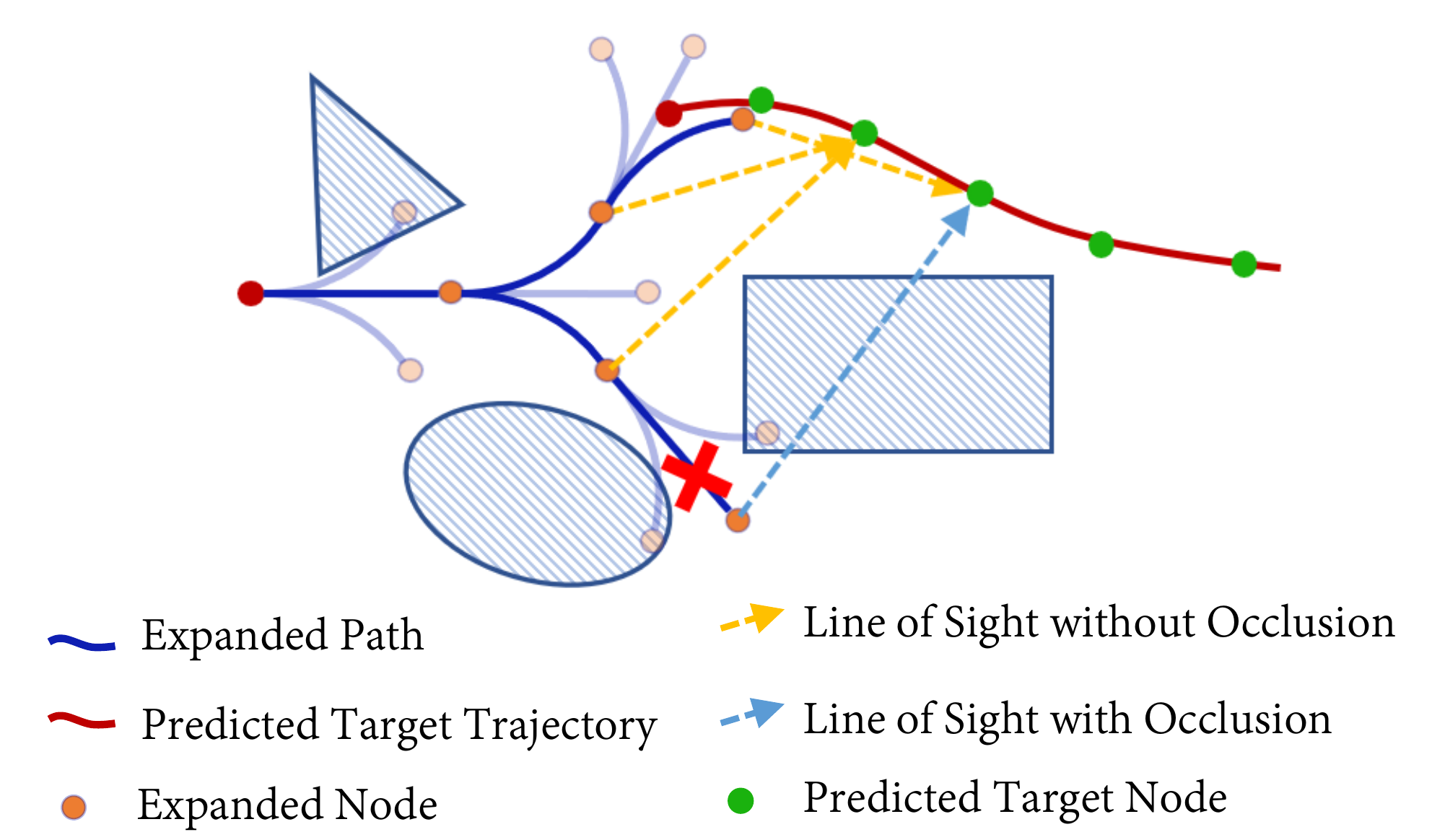}
		\captionsetup{font={small}}
		\caption{
			Illustration of the kinodynamic occlusion-avoid searching method. The expanded node will be rejected if the connection line between the node and the target is occluded, such as the red cross.
		}
		\label{pic:front-end}
		\vspace{-0.6cm}
	\end{figure}
	
	\subsection{Implementation Details}
	For the front-end, we use the kinodynamic occlusion-avoid searching method based on hybrid-state A* algorithm \cite{dolgov2010path}. It uses motion primitives to expand nodes and evaluates them for a safe and dynamically feasible trajectory. To guarantee the topological equivalence between the planning path and the target's predicted path, we design a feasibility check function that prevents occlusion between the robot and the target, as Fig.\ref{pic:front-end} is shown.

	For the back-end, we apply our visibility-aware joint optimization method. Furthermore, a safe tracking cost $\omega_v \cdot J_v$ is added to keep the soon-to-arrive environment known. This cost tends to restrict the velocity direction into the quadrotor's FOV:
	
	\begin{equation}
		J_{v}=\sum_{k=1}^{N_c-2} g\left(  (\psi_{v,k}-\psi_{p,k})^2-\psi_{thr}^2 \right),
	\end{equation}
	where $\psi_v$ is the velocity direction that equals to the tangent direction of the trajectory at the point $\mathbf{p}$, and $\psi_{thr}$ is the maximum allowable deviation angle from FOV axis.


	\section{Results}
	\label{sec:results}
	
	\begin{table}[b]
		\vspace{-0.5cm}
		\centering
		\caption{Parameters of Optimization.}
		\setlength{\tabcolsep}{5.0mm}
		\renewcommand\arraystretch{1.2}
		{
			\begin{tabular}{|c|c|c|c|}
				\hline
				$od_{min}$             & $od_{max}$       & $\rho$ & $M$ \\  \hline
				$2.5m $       & $3.5m$             & $0.8$       & $10$ \\  \hline
		\end{tabular}}

		\label{tab:planner_cmp}
		\vspace{-0.1cm}
	\end{table}
	
	\begin{table}[b]
		\centering
		\caption{Parameters for Experiments.}
		\setlength{\tabcolsep}{1.5mm}
		\renewcommand\arraystretch{1.2}
		{
			\begin{tabular}{|c|c|c|}
				\hline
				& \tabincell{c}{Target Quadrotor \\ Max Velocity}       &  \tabincell{c}{Tracking Quadrotor \\ Max Velocity}\\  \hline
				Real-world        & $1.5 m/s$             & $2.5 m/s$        \\  \hline
				Simulation        & $2.5 m/s$           & $5.0 m/s$       \\  \hline
		\end{tabular}}

		\label{tab:planner_cmp2}
		\vspace{-0.2cm}
	\end{table}	
	
	We incorporate our method into a quadrotor tracking system for testing. We set another high-speed quadrotor as the target. Due to the unknown intent of the target, the future position is predicted by history. The quadrotor performs real-time mapping since the environment is unknown. The goal of the tracking system is to follow the target and keep it in FOV. The parameters of optimization are shown in Tab.\ref{tab:planner_cmp}.

	We conduct two cases and general tests in the simulation and deploy the system on a quadrotor for real-world experiments. For simulation, we benchmark Fast-Tracker\cite{han2020fast} for its cutting-edge performance in tracking. Two special cases are conducted to demonstrate the performance of the proposed method. Case.1 and Case.2 present the validity of our method under \textit{OE} and \textit{AO} metric, respectively. In the general tests, the tracking system runs for a long time in several randomly generated complex environments to comprehensively test our method's visibility effect and robustness. We further demonstrate the practical performance of the algorithm through real-world experiments.

	\begin{figure*}[t]
		\vspace{-0.0cm}
		\centering
		\includegraphics[width=1.0\linewidth]{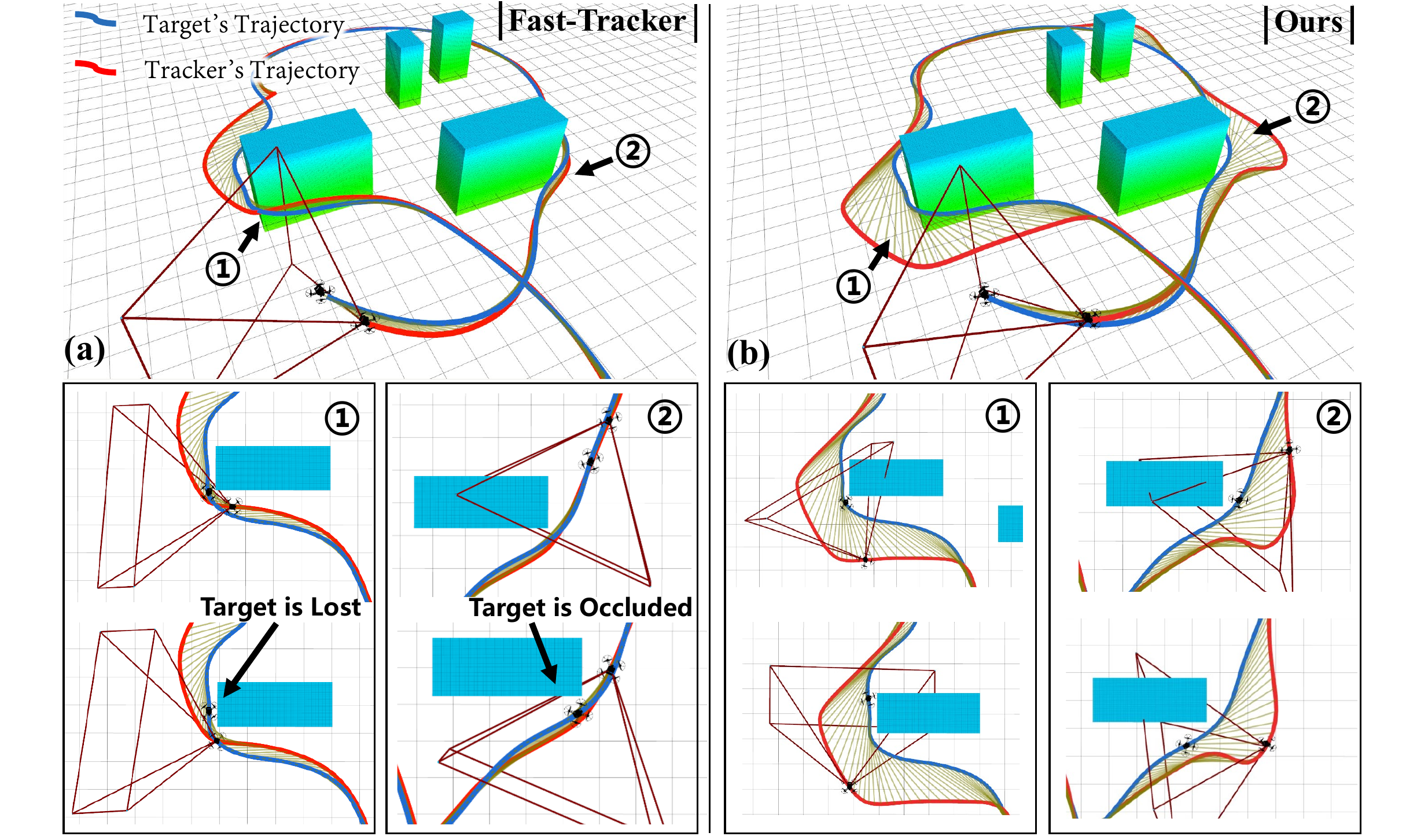}
		\captionsetup{font={small}}
		\caption{
			The simulation results for Case 1. The trajectory of the tracking quadrotor and the target quadrotor are denoted as red and blue, respectively. The yellow lines connect their positions at the corresponding time. 
			\textbf{We use the same color scheme in the following figures
			of simulation experiments.}
			The bottom snapshots show the situation at two corners where occlusion can easily occur. (a): the result of Fast-Tracker. (b): the result of our method.
		}
		\label{pic:EO}
		\vspace{-0.9cm}
	\end{figure*}

	\subsection{Real-World Experiments}	
	\label{sec:experiment} 
	
	Real-world experiments are presented on the same quadrotor platform of \cite{zhou2020ego}, which is localized by a robust visual-inertial state estimator \cite{qin2018vins}. The tracking quadrotor is equipped with an Intel Realsense D435 that has FOV (H × V) = 86° × 57° and an onboard computer DJI Manifold 2C. 
 	The target is an autonomous quadrotor as well broadcasting its location to avoid the impact of the identification, since our proposed method mainly focuses on visibility-aware trajectory generation.
 	In this experiment, the max velocities of the target and the tracking quadrotor are set as Tab.\ref{tab:planner_cmp2}.
	
	In the experiment scene, the target swerves behind an obstacle, which easily causes target loss. As Fig.\ref{pic:snapshot11} shows, to avoid the line of sight toward the target being occluded, the tracking quadrotor adjusts its trajectory to keep the confident FOV obstacle-free. According to the tracking quadrotor camera data, the target is in the the FOV throughout the whole experiment. Furthermore, the tracking trajectory keeps smooth and dynamic feasible. Snapshots and visualization of the experiment can be found in Fig.\ref{pic:snapshot11}. We refer readers to the video for more information\footnote{https://www.youtube.com/watch?v=PhhrOBx54YY}.

	\subsection{Simulation and Benchmark Comparisons}
	\label{sec:benchmark}	
	We benchmark our method with Fast-Tracker presented in the \cite{han2020fast} in simulation. In Fast-Tracker, they searches for a safe tracking trajectory heuristically by a target informed kinodynamic searching method as the front-end. The back-end optimizer then refines the trajectory into a local spatial-temporal optima trajectory\cite{wang2020generating}.
	To compare the tracking performance fairly, we set both of them the same target motion prediction and an appropriate FOV (H × V) = 80° × 65°. 
	Furthermore, we define the target is lost when occlusion occurs or it is out of the FOV.

	\subsubsection{Case 1}
	\textit{OE} metric is validated. In this scene, the target makes sharp turns behind obstacles.
	Fast-Tracker never takes occlusion into account, which causes that the target is easily obstructed by obstacles as shown in Fig.\ref{pic:EO}(a). 
	In contrast, considering \textit{OE}, the planner with our method generates a trajectory that prevents the line of sight towards the target from occlusion, as shown in Fig.\ref{pic:EO}(b). 
	
	\begin{figure}[t]
		\vspace{0.3cm}
		\centering
		\includegraphics[width=1\linewidth]{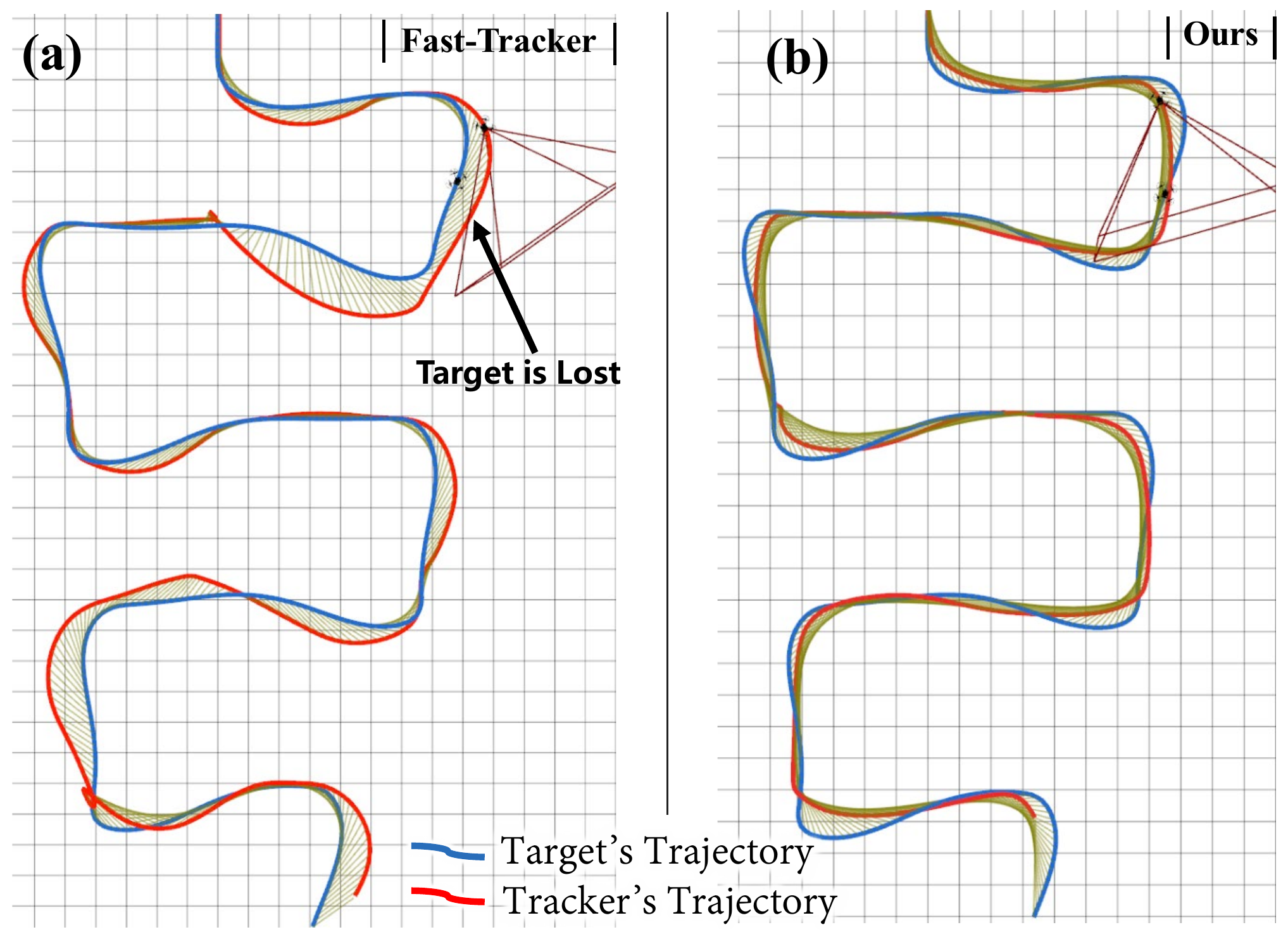}
		\captionsetup{font={small}}
		\caption{
			The test environment of Case 2 for \textit{AO} and \textit{DO}. (a): the result of Fast-Tracker. (b): the result of our method.
		}
		\label{pic:AO_env}
		\vspace{-0.3cm}
	\end{figure}
	\begin{figure}[t]
		\vspace{0.4cm}
		\centering
		\includegraphics[width=1\linewidth]{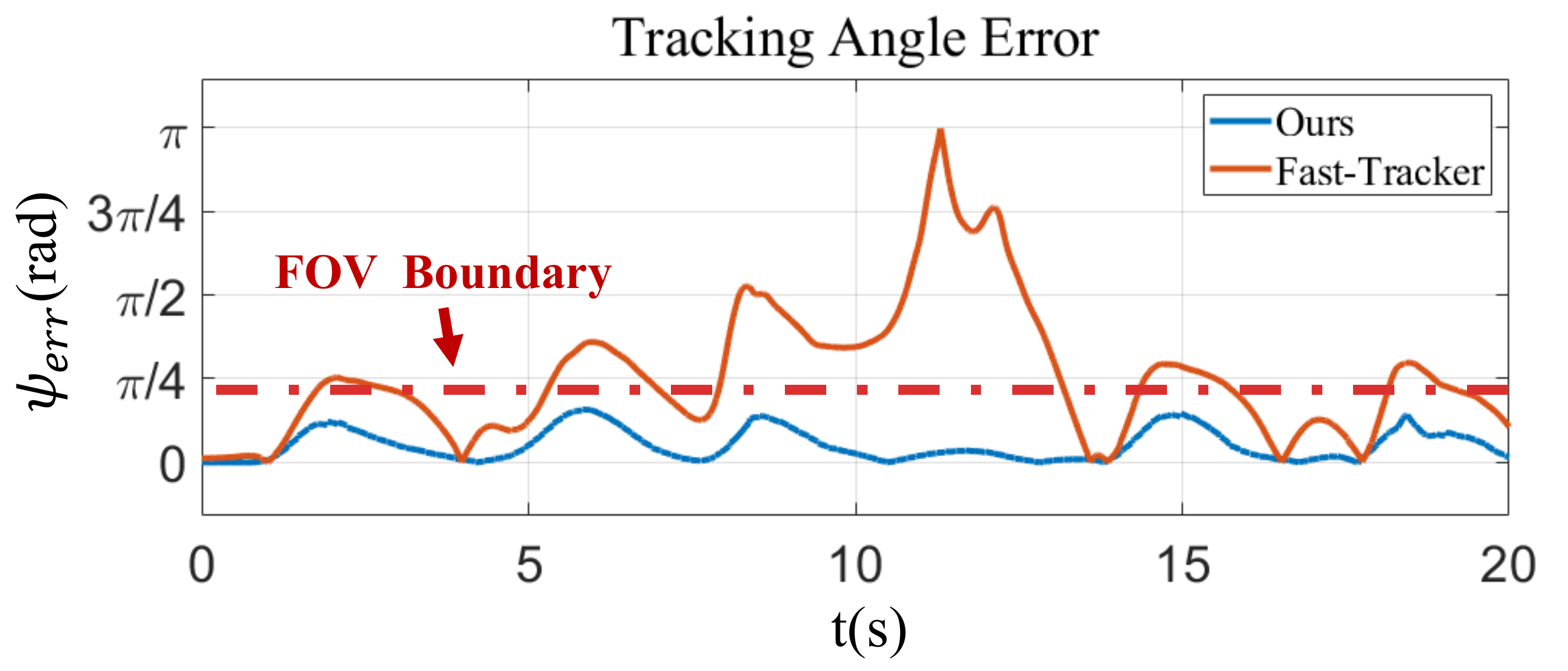}
		\captionsetup{font={small}}
		\caption{
			Comparison of the tracking angle error . The tracking angle error of our method is remarkably less than Fast-Tracker. For our method, $\psi_{err}$ is below FOV boundary for the whole time. In contrast,  in Fast-Tracker, the target is always out of FOV bounds, which leads to target loss.
		}
		\label{pic:ao}
		\vspace{-0.2cm}
	\end{figure}

	\subsubsection{Case 2}
	\textit{AO, DO} metric are validated. In this scene, the target makes large turns that are always more than $90 ^{\circ}$.
	Omitting a moderate distance and yaw angle to observe the target, Fast-Tracker loses it easily when turning, as shown in Fig.\ref{pic:AO_env}. 
	In contrast, taking AO and DO into account, our method jointly optimizes position and yaw, which makes the quadrotor quickly turn toward the target even if there are many big turns.
	For comparision, we define tracking angle error $\psi_{err} = | \psi_p - \psi_{best} |$.
	Fig.\ref{pic:ao} shows the comparison of $\psi_{err}$ in the whole time. The FOV boundary equals half of the FOV angle. Consequently, our method ensures the target in FOV. However, Fast-Tracker almost has the target out of FOV at every turn.
	
\begin{figure}[t]
		\vspace{0.3cm}
		\centering
		\includegraphics[width=1\linewidth]{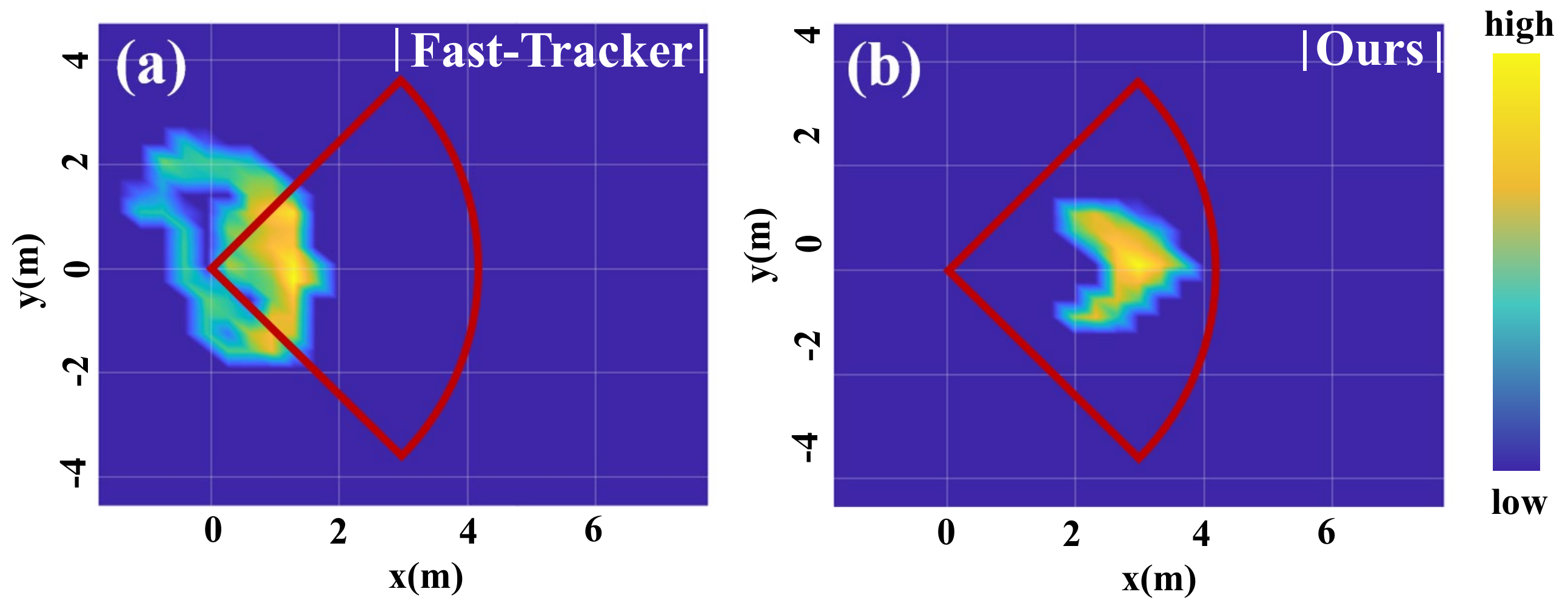}
		\captionsetup{font={small}}
		\caption{
			Heat map comparison. The heat map shows the distribution of the target positions relative to the tracking quadrotor on x-y plane. The red sector represents the FOV of the tracking quadrotor. The closer it is to yellow, the more frequently the target appears. 
			Our approach always restricts the target to the central area. (a): heat map of Fast-Tracker. (b): heat map of our method.
		}
		\label{pic:allheat}
		\vspace{0.4cm}
	\end{figure}

	Furthermore, as is shown in Fig.\ref{pic:allheat}, we count the target positions projected to x-y plane in the tracking quadrotor's FOV. The heat map shows the distribution of the target positions, relative to the tracking quadrotor. For better visualization, we replace the count number $N$ in each heat map grid by $log(N+1)$.
	
	Consequently, our method keeps the target in a moderate area of FOV, which is conducive for observation, while Fast-Tracker always loses the target.

	\subsubsection{General Test}
\begin{figure}[t]
		\vspace{-0.2cm}
		\centering
		\includegraphics[width=1\linewidth]{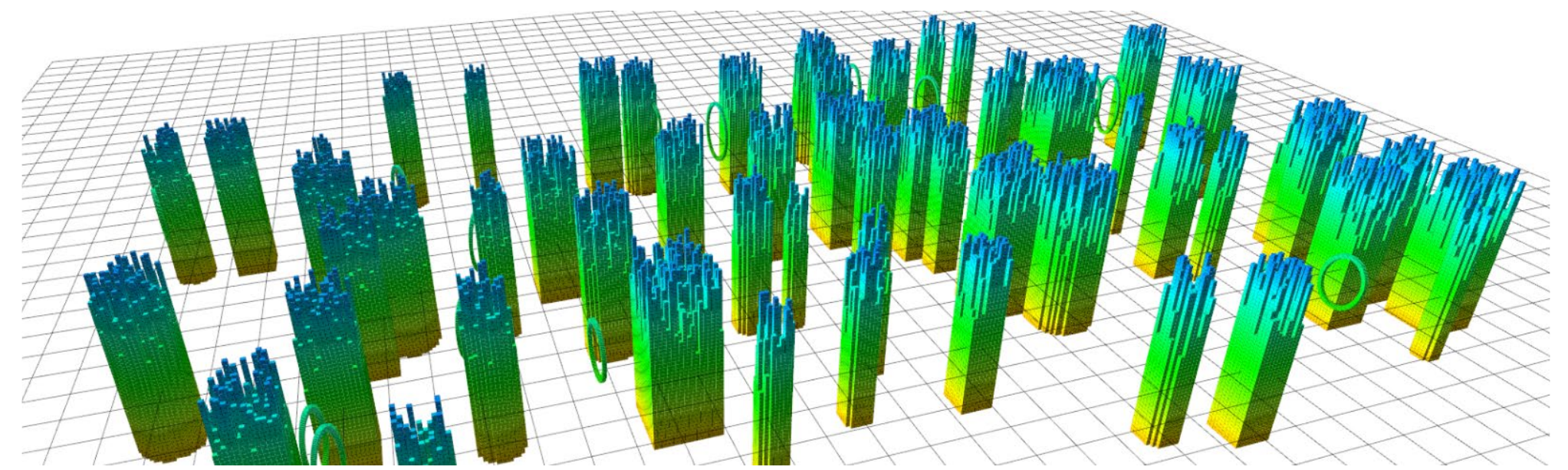}
		\captionsetup{font={small}}
		\caption{
			The random environment in general test.
		}
		\label{pic:test_env}
		\vspace{-1.6cm}
	\end{figure}

	To prove that our method is generic, we compare both methods in an environment generated with randomly deployed obstacles as shown in Fig.\ref{pic:test_env}.

	In a variety of applications, we may be incapable to obtain the target location directly, which requires the quadrotor to identify the target’s location. 
	As a result, we define the tracking mission as a failure when the tracking quadrotor loses the target. 
	In tests, the tracking quadrotor follows the target moving along a random trajectory that lasts for 100s. 
	For comparison, we further define failure time $T_{fast}$ and $T_{our}$ are the time when the mission fails. 
	The result is shown in Tab.\ref{tab:fail_time}. $V_{max}$ is the max velocity  set to the target, and we set double $V_{max}$ to the tracking quadrotor. The failure time are compared in 10 tracking missions for each scenario. Consequently, in general environment, our method maintains a higher success rate, while Fast-Tracker fails easily.


	\begin{table}[h]
		\vspace{-0.2cm}
		\centering
		\caption{Comparision of Failure Time in General Tests.}
		\setlength{\tabcolsep}{5.0mm}
		\renewcommand\arraystretch{1.2}
		{
			\begin{tabular}{|c|c|c|c|}
				\hline
				  $V_{max}$ $(m/s)$            & 0.5      & 1.5 & 2.5 \\  \hline
				$T_{fast}$   $(s)$    & 39.1             & 66.25       & 11.0 \\  \hline
				$T_{our}$    $(s)$   & 100             & 100       & 85.7 \\  \hline
		\end{tabular}}

		\label{tab:fail_time}
		\vspace{-0.1cm}
	\end{table}

	\section{Conclusion}
	\label{sec:conclusion}
	\vspace{-0.14cm}
	In this paper, we summarize  three metrics \textit{DO, AO, OE} to evaluate the visibility according to practical experience. Under these metrics, we design differentiable visibility costs and propose a general visibility-aware trajectory optimization method. An joint optimizer is proposed to adjust the position and yaw simultaneously. To validate the visibility effect of our method, we integrate the optimizer into the back-end of a quadrotor tracking system and benchmark it against a state-of-the-art tracking planner. Simulation comparisons and real-world experiments validate that it is robust and efficient.
	
	In the future, our visibility-aware trajectory optimization method will be extended to active slam and exploration areas. As for aerial tracking, we will continue to improve the prediction method to make it a more robust system.

	\begin{figure*}
		\centering
		\includegraphics[width=0.95\linewidth]{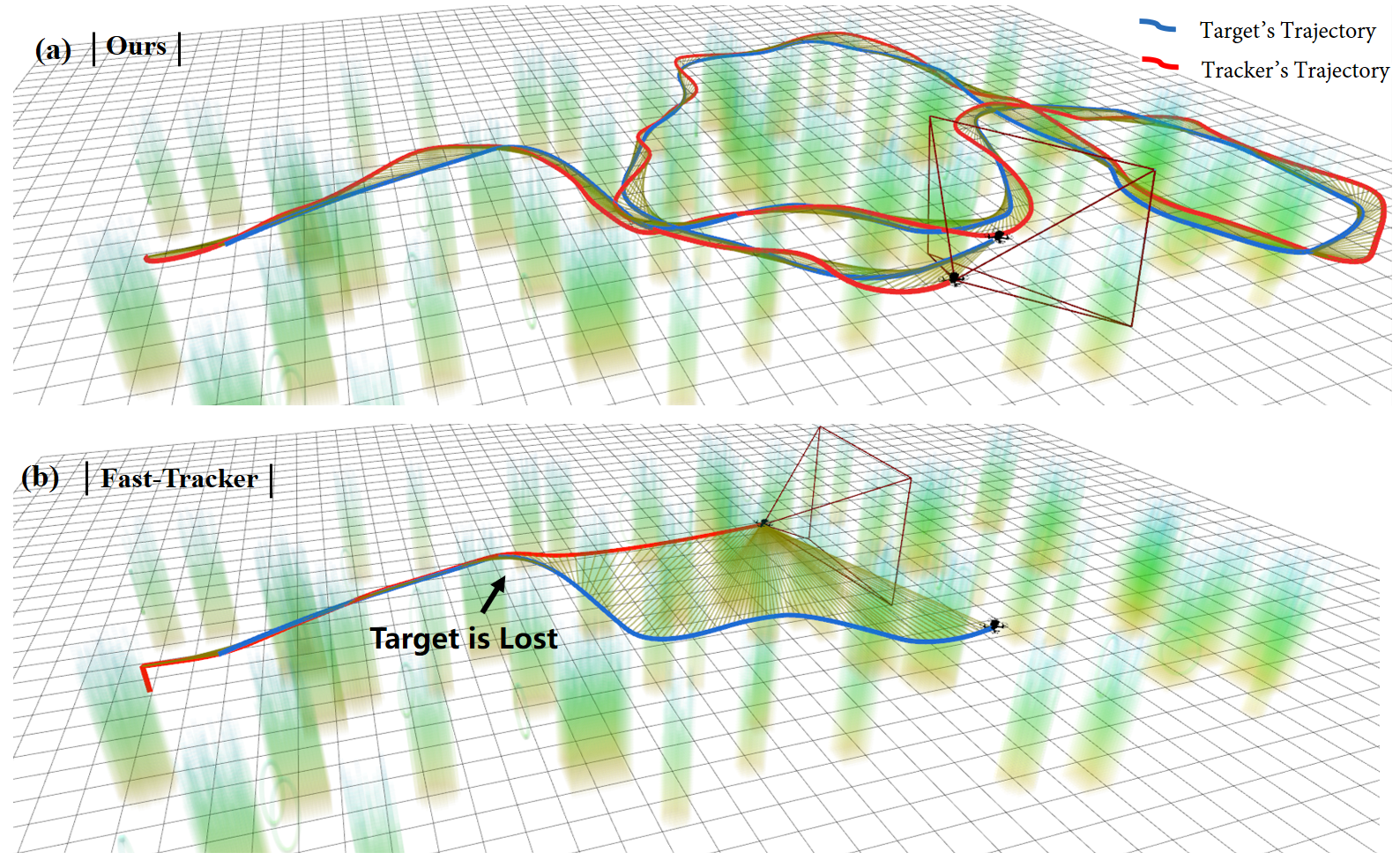}
		\captionsetup{font={small}}
		\caption{
			 Comparision of the trajectories in one of the general tests. (a): the result of our method. (b): the result of Fast-Tracker.
		}
		\label{pic:complex}
		\vspace{0.0cm}
	\end{figure*}

	\newlength{\bibitemsep}\setlength{\bibitemsep}{0.0\baselineskip}
	\newlength{\bibparskip}\setlength{\bibparskip}{0pt}
	\let\oldthebibliography\thebibliography
	\renewcommand\thebibliography[1]{%
		\oldthebibliography{#1}%
		\setlength{\parskip}{\bibitemsep}%
		\setlength{\itemsep}{\bibparskip}%
	}
	\bibliography{iros2021_wxx}
\end{document}